%% file: aaai21.tex
\def\year{2021}\relax
\documentclass[letterpaper]{article} 
\usepackage{aaai21}  
\usepackage{soul}
\usepackage[utf8]{inputenc}
\usepackage{booktabs}

\input{math_commands.tex}

\usepackage[T1]{fontenc}    
\usepackage{nicefrac}       
\usepackage{microtype}      
\usepackage{amsmath}
\usepackage{amsthm}
\usepackage{bm, bbm}
\usepackage{tikz}
\usepackage{multirow}
\usepackage{amsfonts,amssymb}
\usepackage{algorithm,algorithmicx}
\usepackage{subfigure}
 \usepackage{mathrsfs}
\usepackage[noend]{algpseudocode}
\usepackage{times}  
\usepackage{helvet} 
\usepackage{courier}  
\usepackage[hyphens]{url}  
\usepackage{graphicx} 
\graphicspath{{images/}}
\usepackage{natbib}  
\usepackage{caption} 
\newcommand\mypara[1]{\noindent\textbf{#1}}

\theoremstyle{plain}
\newtheorem{assumption}{Assumption}
\newtheorem{prop}{Proposition}

\title{Attack-Resistant Federated Learning with Residual-based Reweighting}
\author {
    Shuhao Fu\textsuperscript{\rm 1}\hspace{5mm}
    Chulin Xie\textsuperscript{\rm 2}\hspace{5mm}
    Bo Li\textsuperscript{\rm 3}\hspace{5mm}
    Qifeng Chen\textsuperscript{\rm 1}\\
}
\affiliations {
    \textsuperscript{\rm 1}HKUST, 
    \textsuperscript{\rm 2}Zhejiang University,     \textsuperscript{\rm 3}UIUC
}
\begin{document}

\maketitle

\begin{abstract}
Federated learning has a variety of applications in multiple domains by utilizing private training data stored on different devices. However, the aggregation process in federated learning is highly vulnerable to adversarial attacks so that the global model may behave abnormally under attacks. To tackle this challenge, we present a novel aggregation algorithm with residual-based reweighting to defend federated learning. Our aggregation algorithm combines repeated median regression with the reweighting scheme in iteratively reweighted least squares. Our experiments show that our aggregation algorithm outperforms other alternative algorithms in the presence of label-flipping and backdoor attacks. We also provide theoretical analysis for our aggregation algorithm.

\end{abstract}

\section{Introduction}

Federated learning is a machine learning methodology for training a global model with decentralized data stored on multiple or millions of devices \citep{McMahan2017}. In federated learning, private data is stored locally in isolated devices and will not be revealed to other parties during training. 
Federated learning can enable numerous real-world machine learning applications by utilizing massive training data that are privacy-sensitive and scattered on different devices \citep{Bonawitz2017}. For instance, multiple hospitals can collaborate to train a global model for classifying diseases using X-ray images without compromising patient privacy. These hospitals may possess X-ray images in different quantities and varieties, resulting in the non-IID (independent and identically distributed) data distribution that is common in federated learning. 

The default federated learning aggregation algorithm \textit{FedAvg} \citep{McMahan2017} that takes the average of locally updated models is vulnerable to various attacks. We find that federated learning suffers from label-flipping and backdoor attacks in our experiments. When a local model is poisoned, the aggregated global model can also be poisoned and fail to behave correctly. 

Mitigating attacks in federated learning or distributed learning has been explored in recent research \citep{Chen2017,Yin2018,Fung2018,Blanchard2017}. Although the median or trimmed mean aggregation algorithms \citep{Yin2018} may seem plausible in distributed learning, their performance degrades in federated learning when data is non-IID. FoolsGold \citep{Fung2018} solves non-IID problem by identifying participants with a similar objective as attackers, but this strategy may not work when some harmless participants have similar local data.
To make federated learning more attack-resistant, we develop an aggregation algorithm that is robust against label-flipping and backdoor attacks in a general non-IID setting. We derive our aggregation algorithm by adopting the repeated median estimator \citep{Siegel1982} and the reweighting scheme in iteratively reweighted least squares (IRLS) \citep{Holland1977,intro}. We estimate the confidence of each parameter in the local models and then the weight of each local model can be computed by heuristically accumulating all the parameter confidence in each local model.

We compare our proposed algorithm to several baselines by conducting experiments on four datasets, the MNIST dataset \citep{mnist}, CIFAR-10 dataset \citep{cifar}, Amazon Reviews dataset \citep{amazon} and the Lending Club loan dataset \citep{loan}.  Our proposed aggregation significantly mitigates the impact of attacked models in non-IID federated learning and outperforms other baselines in our evaluation. Furthermore, we provide theoretical analysis for our aggregation algorithm.

\begin{figure*}[t]
\centering
\includegraphics[width=1.0\linewidth]{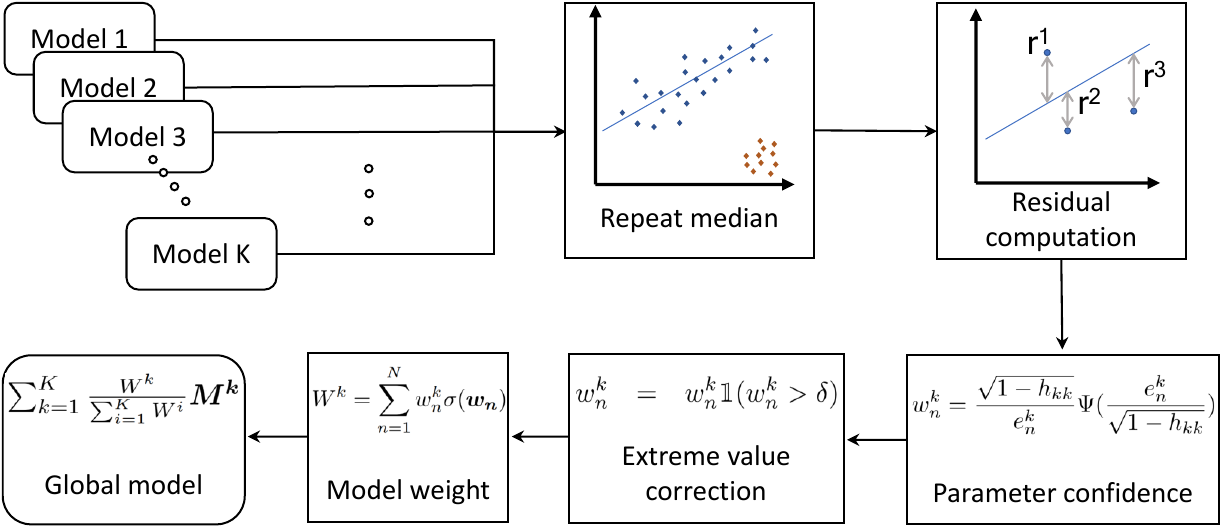}
\caption{The overview of our aggregation algorithm for attack-resistant federated learning.}
\label{fig:overview}
\end{figure*}

\section{Related Work}

\mypara{Adversarial attacks on federated learning.}
Several attacks have been studied against federated learning \citep{Wang2018,Biggio2012,Fung2018,Hayes2019,Hitaj2017,Melis2019}. The label-flipping attack \citep{Biggio2012}  is shown to have great harm to a federated system even with a very small number of attackers \citep{Fung2018}. In this attack, the attacker flips the labels of training data from one class to another class and trains the model accordingly. \cite{Bagdasaryan2018} propose a backdoor attack so that the global model behaves incorrectly on adversarial targeted input. In our work, we mainly focus on defending against label-flipping and backdoor attacks. Note that an attacker can perform any type of attack, such as modifying any model values and training the local model on poisoned data for arbitrary epochs.

\mypara{Robust distributed learning.}
Statistical methods have been studied and applied in robust distributed learning where data is IID \citep{Feng2014,Blanchard2017,Chen2017,Yin2018,Alistarh2018}. The median method and the trimmed mean method \citep{Yin2018} are effective approaches in robust distributed learning but may not be attack-resistant in federated learning where data distribution is non-IID. To tackle the challenge in robust federated learning, we propose a reweighted aggregation algorithm that dynamically assigns weights to the local model based on the residual to a regression line estimated by the repeated median estimator \citep{Siegel1982}.

\mypara{Defending federated learning.}
Recently, some researchers have proposed some defense strategies for robust federated learning \citep{Fung2018,Blanchard2017,Li2019}. FoolsGold \citep{Fung2018} is a defense mechanism against Sybil attacks by adjusting the learning rates of local models based on contribution similarity. The algorithm identifies grouped actions as Sybil attacks and promotes the diversity of local model update. \cite{Gu2018} proposed a model, CalTrain, that represents data with fingerprints to identify poisoned data and models. \cite{Konstantinov2019} propose to maintain a small reference dataset to justify the quality and accountability of models. While this method is effective, it requires a lot of time to evaluate each model in every single round. Our algorithm does not need an additional reference dataset before or after each aggregation process. On the other hand, some researchers proposed to improve the privacy preservation of federated learning \citep{Bonawitz2017,Geyer2017,Truex2018,Thakkar2019}. \cite{Bonawitz2017} propose a privacy-preserving protocol for model aggregation in federated learning. \cite{Geyer2017} introduce differential privacy into federated learning. Instead of enhancing privacy preservation, we focus on the robustness of federated learning so that the global model should behave correctly even when there is a large portion of malicious participants.

\section{Our Algorithm}

In federated learning, there are multiple rounds of communication between participants and a central server for learning a global model. In each round, the global model is shared among the $K$ participants, and a local model on each device is trained on its local private data with the shared global model as initialization. Then all the $K$ local models are sent to the central server to update the global model with an aggregation algorithm. Suppose the participant $k$ has a local model $\bm{M^{(k)}}$, and we can update the global model $\bm{M_{global}}$ by taking the aggregation algorithm of all the $K$ local models.

\subsection{Aggregation Algorithm}
Median is a robust estimator widely used in statistics. However, when the data distribution is non-IID, median often neglects underrepresented data by merely taking a single median value. Hence, in our aggregation algorithm, the global model is designed to be a reweighted average of all the local models where the model weights are estimated robustly.

Algorithm \ref{algorithm:1} summarizes our aggregation algorithm, and a detailed step-by-step description is provided below. Inspired by the reweighting scheme in IRLS \citep{intro}, we reweight each parameter by its vertical distance (residual) to a robust regression line. Then the weight of each local model is computed by accumulating the parameter confidence in each local model.

Let \(y_n^{(k)}\) be the $n$-th parameter of the $k$-th local model.
We use \(\bm{y_n}\) to indicate the list of $n$-th parameters in all the local models.
Let \(\bm{x_n}\) be the indices of \(\bm{y_n}\) sorted in an ascending order. Then \(\bm{(x_n,  y_n)}\) is a point set in 2D with increasing values in the $y$ direction.

\mypara{Repeated median.}
We use the repeated median estimator \citep{Siegel1982} to estimate a linear regression line $y = \beta_{n0} + \beta_{n1}x$. The slope \(\beta_{n1}\) and intercept \(\beta_{n0}\) are estimated as follow,
\begin{eqnarray}
\beta_{n1} &=& \underset{i}{\mathrm{median}}\  \underset{i\neq j}{\mathrm{median}}\ \frac{y_n^{(j)} - y_n^{(i)}}{x_n^{(j)} - x_n^{(i)}}\\
\beta_{n0} &=& \underset{i}{\mathrm{median}}\  \underset{i\neq j}{\mathrm{median}}\ \frac{x_n^{(j)} y_n^{(i)} - x_n^{(i)} y_n^{(j)}}{x_n^{(j)} - x_n^{(i)}} 
\end{eqnarray}
where $i, j \in \{1, 2, ..., K\}$.

\mypara{Residual computation.}
We can calculate the residuals of the $n$-th parameters in all the local models:
\[\bm{r_n} = \bm{y_n} - \beta_{n0} - \beta_{n1} \bm{x_n}.\]
Since \(\bm{r_n}\) can be very different in magnitude for different parameters, we can normalize \(\bm{r_n}\) similar to the reweighting scheme in IRLS \citep{intro}:
\begin{eqnarray}
\tau_n &=& \gamma \widetilde{|r_n|} (1 + \frac{5}{K-1}),\\ 
\widetilde{|r_n|} &=& \mathrm{median} (|\bm{r_n}|).
\end{eqnarray}
where $\gamma$ is a constant.  We set $\gamma = 1.48$ recommended by Wilcox et al. \citep{intro}. Then the normalized residuals become
\begin{equation}
e^{(k)}_n = \frac{r^{(k)}_n}{\tau_n}.
\end{equation}

\begin{algorithm}[t!]
\caption{Our aggregation algorithm}\label{euclid}
    \hspace*{\algorithmicindent} \textbf{Input:} Models $\bm{M^{(1)}}, \bm{M^{(2)}}, ..., \bm{M^{(K)}}$, with parameters $y_n^{(1)}, y_n^{(2)}, ..., y_n^{(K)}$ \\
    \hspace*{\algorithmicindent} \textbf{Output:} The global model $\bm{M_{global}}$ 
\begin{algorithmic}[1]
\For{$n$-th parameter where $ n = 1 \to N \text{, and let }\bm{y_n} = [y_n^{(1)}, y_n^{(2)}, ..., y_n^{(K)}]^T$}
\State $\bm{x_n} \gets $ indices of $\bm{y_n}$ sorted in an ascending order
\State $\beta_{n0}$, $\beta_{n1} \gets $  $RepeatedMedian(\text{$\bm{x_n}$}, \text{$\bm{y_n}$})$ \Comment{get robust line estimation}
\State $\bm{r_n} \gets \bm{y_n} - \beta_{n0} - \beta_{n1} \bm{x_n}$ \Comment{compute residual}
\State $\widetilde{|r_n|} \gets Median|\bm{r_n}|$
\State $\tau_n \gets \gamma \widetilde{|r_n|}(1 + \frac{5}{K-1})$ \Comment{normalize residuals}
\State $\bm{e_n} \gets \frac{\bm{r_n}}{\tau_n}$
\State $\bm{H_n} \gets \bm{x_n}(\bm{x_n}^T\bm{x_n})^{-1}\bm{x_n}^T $\Comment{compute Hat matrix}
\State $\bm{w_n} \gets \frac{\sqrt{\bm{1}-diag(\bm{H_n})}}{\bm{e_n}}\Psi(\frac{\bm{e_n}}{\sqrt{\bm{1}-diag(\bm{H_n})}})$ \Comment{compute parameter confidence}
\State $\bm{y_n}, \bm{w_n} \gets $ CorrectExtremeValue($\bm{y_n}, \bm{w_n}$)
\State $\bm{w_n} \gets \bm{w_n} \sigma(\bm{w_n})$ \Comment{reweight confidence by its standard deviation}
\EndFor
\For{each $k = 1 \to K$}
\State ${W^{(k)}} \gets \sum_{n=1}^{N}\bm{w_n^{(k)}}$ \Comment{accumulate weights}
\EndFor
\State $\bm{M_{global}} \gets \sum_{k=1}^{K}\frac{W^{(k)}}{\sum^K_{i=1}W^{(i)}}\bm{M^{(k)}}$ 
\end{algorithmic}
\label{algorithm:1}
\end{algorithm}

\begin{figure}
  \centering
  \includegraphics[width=1\linewidth]{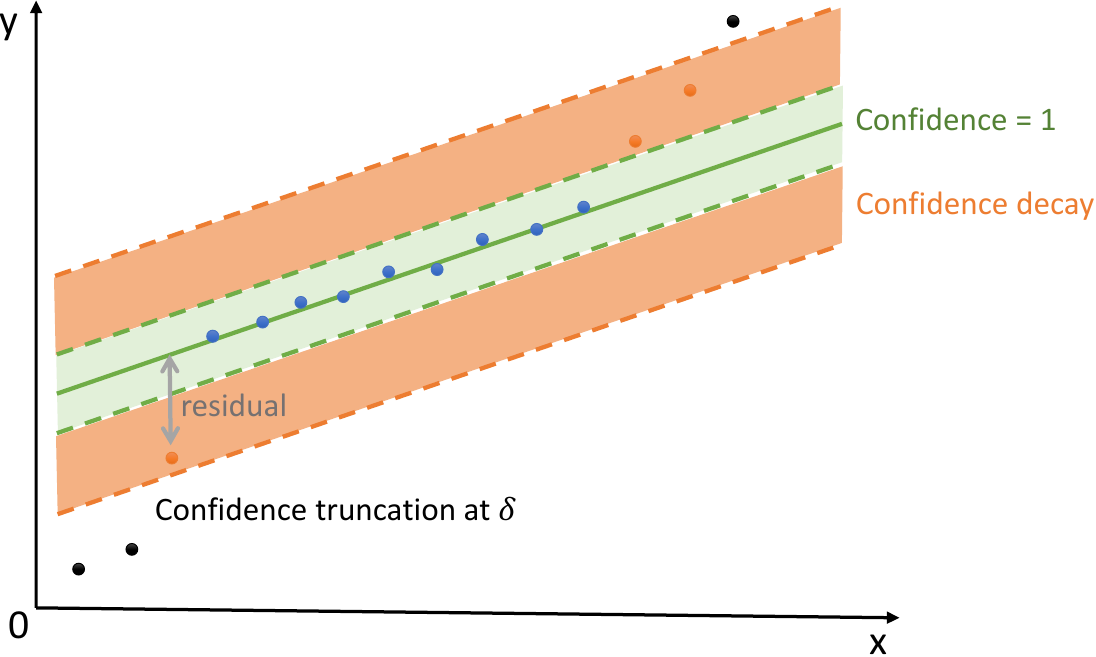}
  \caption{Parameter confidence assignment based on the residual which is the distance from a point to the regression line. In the green area, the parameter confidence is 1; in the orange area, the confidence decays from 1 to $\delta$; in other areas, the confidence is set to 0.}
  \label{fig:residual}
\end{figure}

\mypara{Parameter confidence.}
After obtaining the normalized residuals, the parameter confidence can be determined accordingly \citep{intro}:
\begin{equation}
w^{(k)}_n = \frac{\sqrt{1-h_{kk}}}{e^{(k)}_n}\Psi(\frac{e^{(k)}_n}{\sqrt{1-h_{kk}}}),
\end{equation}
where $w^{(k)}_n$ is the confidence of the $n$-th parameter in $M^{(k)}$, \(\Psi(x) = max\{-Z, min(Z, x)\}\) with \(Z = \lambda\sqrt{2 / K}\) and \(\lambda\) is a hyperparameter. We find $\lambda=2$ is already very robust in all our experiments. \(\Psi\) here acts as a trusted interval and we can expand or shrink the interval by tuning \(\lambda\). \(h_{kk}\) is the k-th diagonal element of matrix in $H_n$:
\begin{equation}
\bm{H_n} = \bm{x_n}(\bm{x_n}^T\bm{x_n})^{-1}\bm{x_n}^T,
\end{equation} where \(\bm{x_n} = [x^{(1)}_n\ x^{(2)}_n\ \hdots\ x^{(K)}_n]^T\).

\mypara{Extreme value correction.}
Extremely large values, even multiplied with a small weight in model aggregation, can damage the global model. We address this issue by involving a a threshold $\delta$. If a parameter has a confidence value lower than \(\delta\), then it should be corrected to the robust line estimation as follows,
\begin{eqnarray}
&w^{(k)}_n = w^{(k)}_n\mathbbm{1}(w^{(k)}_n > \delta),\\
&y^{(k)}_n = y^{(k)}_n\mathbbm{1}(w^{(k)}_n > \delta) + (\beta_{n0} + \beta_{n1} x^{(k)}_n)\mathbbm{1}(w^{(k)}_n \leq \delta).
\end{eqnarray}

\begin{figure*}[t]
\begin{minipage}[t]{0.485\linewidth}
\centering
\includegraphics[width=6.5cm]{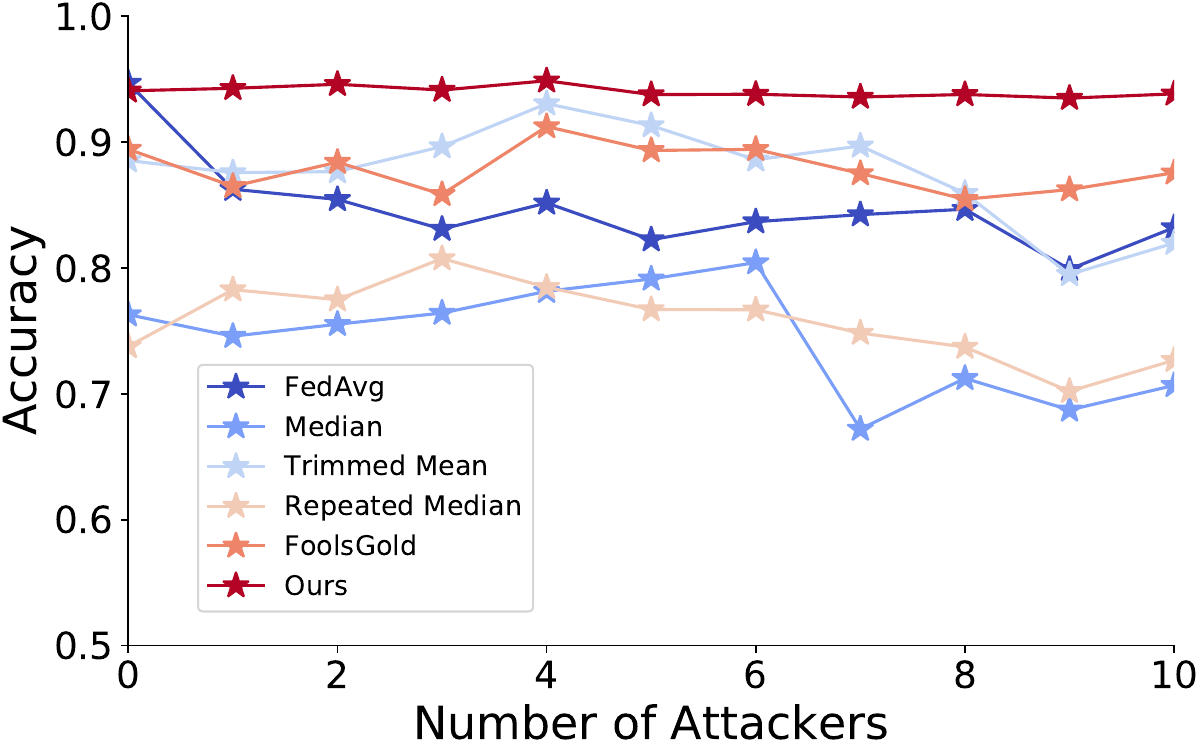}
\caption{Results of label-flipping attacks on the MNIST dataset. The number of participants is 100, within which 0 to 10 of them are attackers.}
\label{fig:mnist_strong}
\end{minipage}
\hspace{3mm}
\begin{minipage}[t]{0.485\linewidth}
\centering
\includegraphics[width=6.5cm]{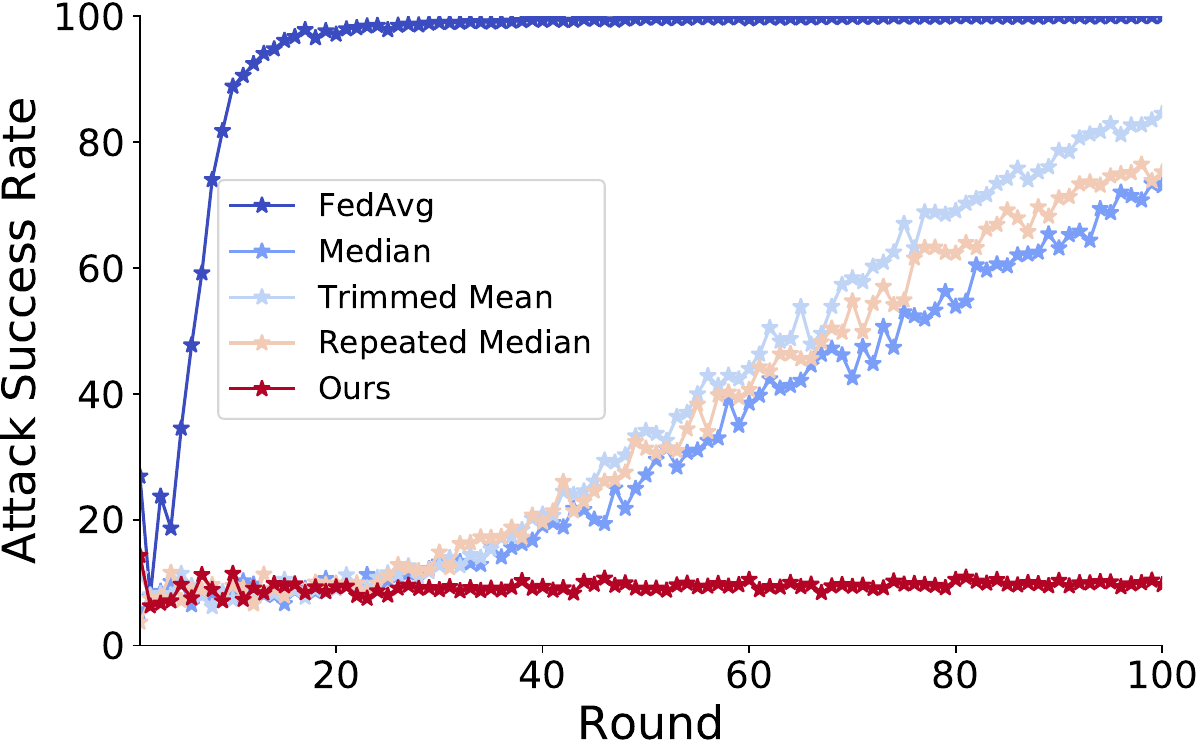}
\caption{Result of backdoor attack success rate on CIFAR-10 under different aggregation algorithms. Ours outperforms other baselines.}
\label{fig: Attack success rate in CIFAR under different aggregation algorithms}
\end{minipage}
\end{figure*}

\mypara{Model weight.}
To obtain the weight of each local model, we can simply aggregate the parameter confidence in the local model but this is not ideal. Imagine an attacker trains a model honestly, but then alters only 10\% of the parameters to some extremely large values. This adversary model still receives about 90\% of the parameter confidence. To address this problem, we measure the importance of a parameter by the standard deviation of $\bm{w_n}$. A confidence assignment with a large standard deviation indicates a great disagreement among this parameter in all models and should be more critical when being accumulated towards model weights:
\begin{equation}
W^{(k)} = \sum_{n=1}^{N}w_n^{(k)} \sigma(\bm{w_n}),
\end{equation}
where \(W^{(k)}\) is the weight for model $k$, $N$ is the number of parameters, \(\bm{w_n}=[w_n^{(1)} w_n^{(2)} \hdots w_n^{(K)}]^T\).

\mypara{Global model.}
Finally, we can obtain the updated global model by
\begin{equation}
    \bm{M_{global}} = \sum_{k=1}^{K}\frac{W^{(k)}}{\sum^K_{i=1}W^{(i)}}\bm{M^{(k)}}.
\end{equation}

\section{Experiments}
We compare our approach with other aggregation algorithms, including FedAvg \citep{McMahan2017}, the coordinate-wise median method \citep{Yin2018}, the coordinate-wise trimmed mean method \citep{Yin2018}, FoolsGold \citep{Fung2018}, and a coordinate-wise repeated median approach we adopt from \citep{Siegel1982}.  We perform experiments on the MNIST handwritten digit dataset \citep{mnist}, the Amazon Reviews dataset \citep{amazon}, the CIFAR-10 dataset \citep{cifar}, and the Lending Club loan dataset \citep{loan}.
We implement attack strategies and defense algorithms in PyTorch \citep{pytorch}.

We use a two-layer convolutional neural network (CNN) for our MNIST experiments. 
With this simple CNN model, our goal is to evaluate different aggregation algorithms for defending federated learning in the presence of attacks. On the CIFAR-10 dataset, we use ResNet-18 \citep{resnet} for image classification. 
The text classification model \textit{FastText} \citep{JoulinGBM16} is adopted for evaluation on the Amazon Reviews dataset. It is a two-layer deep neural network where the first layer is an embedding layer, and the second layer is a fully connected layer. For the Lending Club loan dataset, we use a simple neural network with three fully-connected layers to classify loan status. We use the last two models to demonstrate that our algorithm can be generalized to a natural language processing task and to a real-work financial problem. All the evaluation results are the average of running the same experiments 3 times.

\subsection{Datasets} \label{dataset}

\mypara{MNIST dataset.} The MNIST dataset contains 70,000 real-world handwritten images with digits from 0 to 9. We evaluate different methods by learning a global model on these training images distributed on multiple devices in a non-IID setting with adversarial attacks. 

\mypara{CIFAR-10 dataset.} The CIFAR-10 dataset contains 60,000 natural images in ten object classes. The experimental setup is also non-IID on CIFAR-10.

\mypara{Amazon Reviews dataset.} The Amazon Reviews dataset \citep{amazon} contains product reviews and ratings collected from the Amazon website. Every review is paired with a sentiment rating from 1 to 5.  We categorize comments with rating 1 or 2 as negative and comments with rating 4 or 5 as positive. We discard reviews with rating 3, and we only train a binary classifier. We only use the book reviews from the Kindle Store. 20\% of the reviews are used for testing, while the rest is for training. We obtain a training set of 86,164 reviews and a test set of 13,260 reviews. 

\mypara{Lending Club Loan dataset.} The Lending club dataset LOAN \citep{loan} contains financial information such as credit scores and the number of finance inquiries for loan status prediction (such as ``Current'', ``Late'', or ``Fully Paid''). There are 1,808,534 data samples in 9 types of loan status. We divide them by US states to simulate the federated learning scenarios with non-IID data distribution where each state represents a participant.

\begin{table*}[t!]
\centering
\scalebox{0.9}{
\begin{tabular}{@{}lccccc@{}}
\toprule
Number of attackers & {0} & {1} & {2} & {3} & {4} \\
\midrule
FedAvg \citep{McMahan2017} & 88.96\% & 85.74\% & 82.49\% & 82.35\% & 82.11\%  \\
Median \citep{Yin2018}        & 88.11\% & 87.69\% & 87.15\% & 85.85\% & 82.01\%  \\
Trimmed Mean \citep{Yin2018}  & 88.70\% & 88.52\% & \textbf{87.44\%} & 85.36\% & 82.35\% \\
Repeated Median \citep{Siegel1982} & 88.60\% & 87.76\% & 86.97\% & 85.77\% & 81.82\% \\
FoolsGold \citep{Fung2018} & 9.70\%  & 9.57\%  & 10.72\% & 11.42\% & 9.98\% \\

\midrule
Ours & \textbf{89.17\%} & \textbf{88.60\%} & 86.66\% & \textbf{86.09\%} & \textbf{85.81\%} \\
\bottomrule
\end{tabular}
}
\caption{Results of label-flipping attacks on CIFAR-10 dataset with different numbers of attackers. The total number of participants is 10. }
\label{table:poison}
\end{table*}

\begin{table*}[t!]
\centering
\scalebox{0.9}{
\begin{tabular}{@{}lccccc@{}}
\toprule
Number of attackers & {0} & {1} & {2} & {3} & {4} \\
\midrule
FedAvg \citep{McMahan2017} & \textbf{91.81\%} & 86.91\% & 24.97\% & 12.52\% & 9.78\% \\
Median \citep{Yin2018} & 91.73\% & \textbf{91.87\%} & \textbf{91.79\%} & 91.43\% & 91.17\% \\
Trimmed Mean \citep{Yin2018} & \textbf{91.81\%} & 91.82\% & 91.82\% & 91.49\% & 91.26\% \\
Repeated Median \citep{Siegel1982} & 91.55\% & 88.41\% & 23.22\% & 11.70\% & 9.62\% \\
FoolsGold \citep{Fung2018} & 50.79\% & 49.45\% & 47.44\% & 49.71\% & 49.95\% \\

\midrule
Ours & 91.71\% & 91.79\% & 91.76\% & \textbf{91.67\%} & \textbf{91.38\%} \\
\bottomrule
\end{tabular}
}
\caption{Results of label-flipping attacks on Amazon Reviews dataset with different numbers of attackers. The total number of participants is 10. }
\label{table:poison_amazon}
\end{table*}

\subsection{Results on Label-flipping Attacks} \label{label_flipping}
We evaluate the overall classification performance of different aggregation methods on three datasets under label-flipping attacks, the MNIST dataset  \citep{Biggio2012}, the CIFAR-10 dataset \citep{cifar}, and Amazon Reviews dataset \citep{amazon}. In label-flipping attacks, attackers flip the labels of training examples in the source class to a target class and train their models accordingly. 

In the MNIST experiment, we simulate federated learning with 100 participants, within which 0 to 10 of them are attackers. Each  participant contains images of two random digits. The attackers are chosen to be some participants with images of digit 1 and another random digit since they are flipping the label of 1 to 7. We run 200 synchronization rounds with $\delta$ set to 0.01. In each round of federated learning, each participant is supposed to train the local model for 5 epochs, but the attackers can train for arbitrary epochs. The results are shown in Figure \ref{fig:mnist_strong} where attackers train the models with 5 more epochs to enhance the attacks. Our algorithm outperforms all other methods and is robust when the number of attackers increases. Median methods (Median and Repeated Median) have relatively low accuracy even without any attackers due to their discarding most of the information in model aggregation. Our algorithm, on the other hand, takes the reweighted average of all local models and thus gathers more information in an unsupervised way. We also compare our algorithm with the state-of-the-art algorithm FoolsGold \citep{Fung2018}. Though their algorithm also maintains a low attack success rate, our algorithm is more stable and surpasses FoolsGold by 6\% on average.

For the CIFAR-10 experiment, following \cite{Bagdasaryan2018}, we adopt a Dirichlet distribution \citep{minka2000estimating} with a hyperparameter 0.9 to generate non-IID data distribution for totally 10 participants. The experimental setup is the same for the CIFAR-10 and MNIST experiments under backdoor attacks in Section \ref{result_on_backdoor}. 
The attackers flip the label of ``cat'' to ``dog'' since they are the most similar classes in CIFAR-10. The experiment results can be found in Table \ref{table:poison}.
In this experiment, baseline methods perform fairly well because the data distribution is not immensely non-IID., where each user has all the labels but in different amounts. FoolsGold \citep{Fung2018} fails because they assume that the gradients of honest models are very different because of the non-IID data, and the gradients of outliers are close because they share the same objective. However, it may not always be the case when the data distribution is not so extremely non-IID, such as the CIFAR-10 case. FoolsGold may think these honest users are close and decide they are outliers. Our algorithm, on the other hand, can adapt to extremely non-IID cases, such as the MNIST experiments, where all baseline methods fail, and is also suitable for common non-IID situations such as the CIFAR-10 and the Amazon Review experiments.

In the experiment on Amazon Reviews, there are 10 participants, where 0 to 4 participants are attackers who flip all their labels. Attackers also train for 5 more epochs. We run 10 synchronization rounds in the experiment. The result is summarized in Table \ref{table:poison_amazon}. Our algorithm achieves comparable state-of-the-art results, and our performance does not degrade when there are less than 50\% attackers.

\subsection{Results on Backdoor Attacks}\label{result_on_backdoor}

\begin{figure}
\centering
\begin{tabular}{c@{}c}
\includegraphics[width=1in]{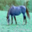} \ \ \ \ \ \ 
&\includegraphics[width=1in]{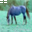}\\
\small{Clean image} & \small{Backdoored image} \\
\end{tabular}
\caption{There is a white color pattern in the left uppe corner of the backdoored image.}
\label{fig: example of backdoored image}
\end{figure}

For pixel-pattern backdoor attacks \citep{gu2017badnets} in federated learning\citep{Bagdasaryan2018}, attackers manipulate their local models so that the learned global model predicts some backdoor target label for any image embedded with certain patterns. 
An example is shown in Figure \ref{fig: example of backdoored image}.
The global model can behave normally for clean data. We choose ``bird'' in CIFAR-10 and ``2'' in MNIST as the backdoor target labels. Similarly, for the preprocessed LOAN dataset that contains 92 features, we manipulate 6 features by assigning certain large values to them and change the labels of manipulated data to ``Does not meet the credit policy. Status:Fully Paid.'' 

The training data is mixed with manipulated data and clean data to fit both the backdoor task and the main task. We compare the performance of aggregation algorithms under two backdoor attack scenarios, which are called the naive approach and the model replacement in \cite{Bagdasaryan2018}. For the naive approach, an attacker poisons its local model and submits the malicious update in every round. For the model replacement, an attacker only poisons in one round to embed some patterns into the global model, so the attacker needs to scale up its malicious update before submission. In our experiment, the malicious participant attacks in round 6 and scales up its update by 100. We run 200 rounds for MNIST and 100 rounds for CIFAR and LOAN.

Table \ref{table:backdoor} summarizes the results of backdoor attacks. 
Our method is the highest in terms of accuracy on MNIST under both backdoor attack scenarios. 
Moreover, on the more challenging CIFAR-10 dataset, our algorithm is the only one that can defend the naive approach backdoor attack.  In Figure \ref{fig: Attack success rate in CIFAR under different aggregation algorithms}, we plot the attack success rates over time under different aggregation algorithms except for FoolsGold because it completely fails in both main and backdoor tasks. Intuitively, backdoor attacks can easily succeed under FedAvg, and other baselines slow down the process but still reach high attack success rate into over 70\% within 100 rounds. Our algorithm effectively defends the attack and remains stable with 9.65\% attack success rate when being attacked continuously for 100 rounds.  
On the LOAN dataset, our method achieves higher accuracy 94.50\% (FedAvg 93.65\%) and 0.00\% attack success rate (FedAvg 99.71\%) under the naive approach attack after 100 rounds. Similarly, our method has 95.06\% accuracy (FedAvg 94.11\%) and 0.00\% attack success rate (FedAvg 98.96\%) under model replacement attacks after 100 rounds. Other baselines also have 0.00\% attack success rate in two attacking scenarios except FoolsGold, whose attack success rate is 99.96\% under the naive approach attack.

\begin{figure}[t]
\centering
\includegraphics[width=0.8\linewidth]{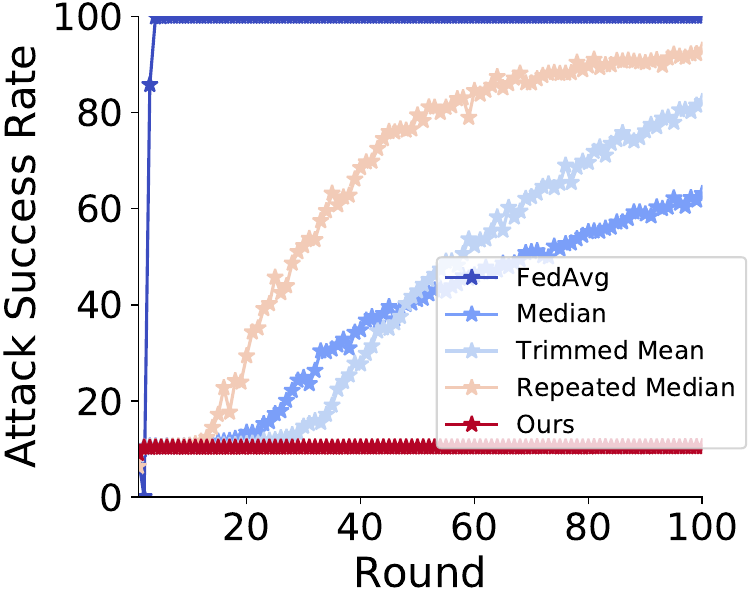}
\caption{Results of backdoor attack success rate on the MNIST dataset with ResNet-18. Our algorithm is still stable while others fail for the complicated model.}
\label{fig:resnet_mnist}
\end{figure}

It is also interesting to notice that all the baselines fail on the CIFAR-10 dataset for the naive backdoor attack approach while they perform fairly well on the MNIST dataset. This is because the model used for CIFAR-10 experiments, ResNet-18, is more complicated and has more parameters ($1000\times$) than the simple CNN used for the MNIST dataset. To explore the effect of complex models, we also train a ResNet-18 model on the MNIST dataset and performed the same backdoor attack on it. The results are shown in Figure \ref{fig:resnet_mnist} except for FoolsGold because it completely fails in both main task (about 10.00\% accuracy) and backdoor task. The results are similar to the same model on the CIFAR-10 dataset. All the baselines fail to defend, but our algorithm remains stable. Besides, most baseline methods (FedAvg, Median, Trimmed Mean, and Repeated Median) are coordinate-wise operations, while our algorithm accumulates a weight for each model rather than each parameter. We believe the model-wise reweighting scheme preserves the structure of the parameters and can perceive higher-level information.

\begin{table*}[t!]
\centering
\scalebox{0.93}{
\begin{tabular}{l@{\hspace{12mm}}ll@{\hspace{6mm}}ll@{\hspace{12mm}}ll@{\hspace{6mm}}ll}
\toprule
\multirow{3}{*}{Dataset} &
\multicolumn{4}{@{\hspace{-12mm}}c}{MNIST} &
\multicolumn{4}{@{\hspace{-3mm}}c}{CIFAR-10} \\
& \multicolumn{2}{l}{Naive approach} & \multicolumn{2}{l}{Model replacement} & \multicolumn{2}{l}{Naive approach} & \multicolumn{2}{l}{Model replacement}  \\
& {Accuracy} & {A.S.R} & {Accuracy} & {A.S.R} & {Accuracy} & {A.S.R} & {Accuracy} & {A.S.R}   \\
\midrule
FedAvg  & \textbf{99.08\%} & 99.71\%  &98.75\% & 17.85\% & 87.44\% & 99.91\% & 69.72\%&  38.59\% \\
Median &  98.91\% & \textbf{10.34\%} &  98.87\% & 10.35\% & 88.58\% & 73.06\% & 87.22\% & 10.01\% \\
Trimmed Mean & 98.97\% & \textbf{10.34\%} & 98.81\% & 10.34\% & 88.38\% & 84.56\% & 87.30\% & 9.85\% \\
Repeated Median  & 98.96\% & 10.36\% & 98.82\% & 10.32\% & 88.22\% & 75.25\% & \textbf{87.57\%} & 9.79\% \\
FoolsGold & 96.20\% & 12.51\%& 97.96\% & \textbf{10.27\%} & 10.00\% &  0.00\% & 10.00\% & 0.00\% \\
\midrule
Our & 98.97\% & 10.35\% & \textbf{98.88\%} & 10.31\% & \textbf{88.89\%} & \textbf{9.65\%} & 87.43\%& \textbf{9.56\%} \\ 
\bottomrule
\end{tabular}
}
\caption{Results of backdoor attacks on MNIST and CIFAR-10. There are 10 participants, 1 of whom is an attacker. We denote the attack success rate as A.S.R. }
\label{table:backdoor}
\end{table*}

\begin{table*}[t!]
\centering
\begin{tabular}{ll|ll|ll|ll|ll}
\toprule
&       
& \multicolumn{2}{l|}{Original}
& \multicolumn{2}{l|}{Median Estimator}    
& \multicolumn{2}{l|}{Theil-Sen Estimator} 
& \multicolumn{2}{l}{Gaussian Weighting} \\ 
\cline{3-10} 
                        &       
& \multicolumn{2}{l|}{Number of attackers} 
& \multicolumn{2}{l|}{Number of attackers} 
& \multicolumn{2}{l|}{Number of Attackers} 
& \multicolumn{2}{l}{Number of Attackers}       \\
$\lambda$ (or $\sigma$ in Gaussian) & Delta 
 & 0                  & 9
 & 0                  & 9                  
 & 0                  & 9                  
 & 0                  & 9                     \\ \hline
1            & 0.01  
& 94.41\%  & 94.54\%  & 94.46\%             & \textbf{95.19\%}            & 93.76\%            & 92.87\%            & 84.32\%               & 92.22\%               \\
1                                   & 0.05  & 93.36\% & 91.15\% & 93.37\%             & 93.79\%            & 94.43\%            & 92.70\%            & 87.33\%               & 90.65\%               \\
1                                   & 0.1   & 86.93\% & 89.39\% & 83.77\%             & 90.93\%            & 92.77\%            & 94.31\%            & 88.31\%               & 89.23\%               \\
1                                   & 0.2   & 84.77\%  & 91.40\% & 70.84\%             & 80.79\%            & 93.28\%            & 93.63\%            & 83.22\%               & 90.28\%               \\ \hline
2                                   & 0.01  & \textbf{94.95\%}  & 94.86\% & 93.34\%             & 94.36\%            & 94.38\%            & 49.28\%            & 91.07\%               & 92.70\%               \\
2                                   & 0.05  & 91.45\%  & 93.14\% & 93.41\%             & 94.86\%            & \textbf{95.62\%}            & 91.65\%            & 90.85\%               & 93.00\%               \\
2                                   & 0.1   & 93.08\%  & 91.84\% & 94.02\%             & 93.48\%            & 92.29\%            & 93.07\%            & 88.61\%               & 93.15\%               \\
2                                   & 0.2  & 86.09\%  & 91.43\%  & 88.84\%             & 92.68\%            & 92.21\%            & 91.70\%            & 90.54\%               & 90.80\%               \\ \hline
3                                   & 0.01 & 93.83\%  & 94.89\%  & 94.67\%             & 94.68\%            & 94.45\%            & 75.83\%            & 92.46\%               & 93.18\%               \\
3                                   & 0.05  & 93.76\%  & \textbf{95.86\%} & 93.67\%             & 94.52\%            & 94.86\%            & \textbf{94.72\%}            & 93.30\%               & \textbf{94.25\%}               \\
3                                   & 0.1  & 94.74\%  & 94.13\%  & 93.11\%             & 91.30\%            & 92.32\%            & 94.70\%            & 92.09\%               & 93.65\%               \\
3                                   & 0.2  & 89.11\%  & 93.25\%  & 93.67\%             & 93.76\%            & 94.00\%            & 93.20\%            & 90.88\%               & 93.26\%               \\ \hline
5                                   & 0.01 & 92.62\%  & 93.77\%  & 93.68\%             & 84.26\%            & 94.69\%            & 93.27\%            & \textbf{94.10\%}               & 93.58\%               \\
5                                   & 0.05 & 94.53\%  & 95.28\%  & 94.23\%             & 94.72\%            & 93.67\%            & 79.91\%            & 92.78\%               & 93.69\%               \\
5                                   & 0.1 & 94.23\%  & 94.47\%   & \textbf{94.88\%}             & 94.69\%            & 94.60\%            & 92.85\%            & 92.81\%               & 93.83\%               \\
5                                   & 0.2 & 92.60\%  & 94.23\%   & 92.90\%             & 93.87\%            & 93.51\%            & 91.41\%            & 91.72\%               & 92.93\%              \\
\bottomrule
\end{tabular}

\caption{The results of the controlled experiments by replacing the linear estimator or the weighting scheme with alternative methods. All the experiments are performed on the MNIST dataset with label-flipping attacks.}
\label{table:alternative}
\end{table*} 

\subsection{Hyperparameters} \label{hyperparameters}
 We have two hyperparameters in our method,$\lambda$ and $\delta$, where $\lambda$ controls the confidence interval, and $\delta$ controls the clipping threshold. We perform a grid search to prove that the hyperparameter selection is robust and efficient. The results can be found in Table \ref{table:alternative}. Basically, a larger $\delta$ means that it has a lower tolerance of differences and thus may eliminate some of the honest models when data is non-IID while increasing $\lambda$ brings insensitivity to the variance of $\delta$. When outliers are easy to be identified (e.g., in IID settings), a large $\lambda$ yields good results with great stability. In the non-IID experiments, however, the models are more diverse, so it is important to find a balance between excluding adversarial models while embracing models trained on different data distributions. In summary, $\lambda$ and $\delta$ are robust in a large range, as shown in Table \ref{table:alternative}.

\subsection{Controlled Experiments}
 Although our algorithm is inspired by the reweighting scheme from IRLS \citep{intro}, some alternative linear estimators and weighting schemes can replace the original ones. In this section, we replace the repeated median estimator in our algorithm with the Theil-Sen estimator or the median estimator and replace the weighting scheme with a zero-mean Gaussian function where smaller residual means larger weight. We tune the $\lambda$ (in the Gaussian weighting it is the variance $\sigma$), and the results are shown in table \ref{table:alternative}. The experiments are performed on the MNIST dataset with label-flipping attacks. The median and Theil-Sen estimators achieve similar results as the Repeated Median estimator, but they can be attacked in a few cases. For example, $\lambda = 5$ and $\delta = 0.01$ for the median estimator and $\lambda = 5$ and $\delta = 0.05$ for the Theil-Sen estimator. The Theil-Sen estimator, especially, only has a relatively low breakdown point than Repeated Median, meaning that it will more easily be broken when the number of attackers increases. The alternative Gaussian weighting scheme is fairly robust against label-flipping attacks. This demonstrates that the robustness comes from our algorithm rather than intricate hyperparameter tuning. 

\section{Theoretical Analysis} \label{sec_proof}
For simplicity, we consider the bound of the error rate for training a single-parameter model on $K$ devices, each storing $S$ IID samples of data. We define the parameter of the local model $i$ as \(\hat{y}^{(i)}\). Suppose that training data points are sampled from some unknown distribution $\mathcal{D}$. Let \(f(y;x)\) be a loss function of the parameter $y \in \mathcal{Y}$ associated with the data point \(x\), where $\mathcal{Y}$ is the parameter space, and $F(y) := \mathbb{E}_{x \in \mathcal{D}}[f(y:x)]$. Our goal is to learn an optimal model defined by the parameter that minimizes the population loss
\begin{equation}\nonumber
\mu =  \arg\min_{y \in \mathcal{Y}} F(y).
\end{equation}
The parameter space $ \mathcal{Y}$ is assumed to be convex and compact with diameter D, i.e., $||y - y^\prime||_2 \leq D$, $\forall y$, $y^\prime \in \mathcal{Y}$.
Suppose that there are \(K\) devices and \(U\) of them are corrupted. We denote the set of adversarial devices as $\mathcal{B}$ and the corruption ratio $\alpha=\frac{U}{K}$. Based on our algorithm, the residuals can be simplified as \((\hat{y}^{(i)} - \tilde{y})\), where \(\tilde{y}\) is the median of \(\{\hat{y}^{(i)}\}\) for \(i = 1, 2, ..., K\). Let $\widetilde{|r|}$ be the median of absolute residuals, i.e., \(\widetilde{|r|} = \mathrm{median}(|\hat{y}^{(i)} - \tilde{y}|)\).
Then, the normalized residual can be expressed as 
\begin{equation}
e^{(i)} = \frac{\hat{y}^{(i)} - \tilde{y}}{\gamma\widetilde{|r|}(1+\frac{5}{K-1})},
\label{eqn:normalized_residual}
\end{equation}
and the parameter confidence is defined as 
\begin{equation}
  z^{(i)} = \left \{
  \begin{aligned}
    & 1, && \text{if}\ \left|e^{(i)}\right| \leq \frac{\sqrt{2}\lambda}{\sqrt{K}} \\
    & \left|\frac{\sqrt{2}\lambda}{\sqrt{K}e^{(i)}}\right|, &&  \text{if}\ \frac{\sqrt{2}\lambda}{\sqrt{K}} < \left|e^{(i)}\right| \leq \left|\frac{\sqrt{2}\lambda}{\sqrt{K}\delta}\right| \\
    & 0, && \text{if}\ \left|e^{(i)}\right| > \left|\frac{\sqrt{2}\lambda}{\sqrt{K}\delta}\right|
  \end{aligned} \right.
  \label{eqn:confidence}
\end{equation} 
Then we will prove that the error of the global model $\bm{M_{global}}:=\frac{1}{\sum^K_{j=1}z^{(j)}}\sum^K_{i=1}z^{(i)} \hat{y}^{(i)}$ is bounded:
\begin{equation} 
\left| \frac{1}{\sum^K_{j=1}z^{(j)}}\sum^K_{i=1}z^{(i)} \hat{y}^{(i)} - \mu \right|= \Tilde{\mathcal{O}}(\frac{1}{\sqrt{S}} + \frac{1}{\sqrt{SK}} + \frac{1}{S} + \frac{1}{\sqrt{K}\delta}).
\label{eqn:proof}
\end{equation}

\begin{proof}
We adopt assumptions 1, 2, 3, and 6, Theorem 8 and Lemma 3 from \cite{Yin2018} here.
We can separate Equation \ref{eqn:proof} into two sets with adversarial participant \(\mathcal{B}\) and normal users \([K] \setminus \mathcal{B}\) where $[K]=\{1,2,\hdots,K\}$. 

\begin{equation}
\resizebox{1.06\hsize}{!}{
$
\begin{aligned}
\left| \frac{1}{\sum^K_{j=1}z^{(j)}}\sum^K_{i=1}z^{(i)} \hat{y}^{(i)} - \mu \right|
& \leq  \frac{1}{\sum^K_{j=1}z^{(j)}}\sum^K_{i=1}z^{(i)} \left| \hat{y}^{(i)} - \mu\right| \\
& \leq \frac{1}{\sum^K_{j=1}z^{(j)}} \left(\sum_{i \in [K] \setminus \mathcal{B}}z^{(i)} \left|\hat{y}^{(i)} - \mu \right| +  \sum_{i \in \mathcal{B}}z^{(i)} \left|\hat{y}^{(i)} - \tilde{y} + \tilde{y} - \mu \right| \right)\\
& \leq \underset{i \in [K] \setminus \mathcal{B}}{\mathrm{max}}\{\left|\hat{y}^{(i)} - \mu \right|\} + \frac{1}{\sum^K_{j=1}z^{(j)}}\sum_{i \in \mathcal{B}}z^{(i)} \left|\hat{y}^{(i)} -\tilde{y} \right| + \left| \tilde{y} - \mu \right|.
\end{aligned}
$
}
\label{eqn:expansion}
\end{equation}

Now we state here the assumptions and theorems from \cite{Yin2018}.
\begin{assumption}[Smoothness of f and F, Assumption 1 from \cite{Yin2018}]\label{as:smoothness} For any $x \in \mathcal{D}$, the derivative of $f(y;x)$ is L-Lipschitz, and the function $f(y;x)$ is L-smooth.
\end{assumption} 

\begin{assumption}[Sub-exponential gradients, Assumption 6 from \cite{Yin2018}]\label{as:subexp} We assume that the derivative of $f(y;x)$ with respect to $y$ is v-sub-exponential.
\end{assumption}

\begin{prop}[Theorem 8 from \cite{Yin2018}]
Define the median of $y^{(i)}$: $\tilde{y} = \mathrm{median}\{y^{(i)}: i \in [K]\}$. Suppose that Assumption \ref{as:smoothness} holds, F(y) is $\lambda_F$-strongly convex. Then, with probability at least $1 - \frac{4}{1+SKLD}$, we have

\begin{equation}
\resizebox{1.\hsize}{!}{
$
\left|\tilde{y} - \mu\right| \leq  \Tilde{\mathcal{O}}\left(\frac{1}{SK} + \frac{1}{\sqrt{S}}\left( \alpha + \sqrt{\frac{\log(1 + SKLD)}{K(1 - \alpha)}} + \frac{1}{\sqrt{S}} \right)\right) = \Tilde{\mathcal{O}}(\frac{\alpha}{\sqrt{S}} + \frac{1}{\sqrt{SK}} + \frac{1}{S})
$
}
\end{equation}
\label{prop:thm8}

\end{prop}

\begin{prop}[Lemma 3 from \cite{Yin2018}]
Suppose that the derivative of the loss function f is i.i.d. v-sub-exponential. Then with a high probability, we have 
\begin{equation}
\resizebox{1.\hsize}{!}{
    $
    \underset{i \in [K] \setminus \mathcal{B}}{\mathrm{max}}\{\left|\hat{y}^{(i)} - \mu \right|\} \leq \frac{2v}{\sqrt{S}}\sqrt{\log(1+SKLD) + \log K} = \Tilde{\mathcal{O}}(\frac{1}{\sqrt{S}})
    $
    }
\end{equation}
\label{prop:lemma3}
\end{prop}

\cite{Yin2018} prove the difference between the median of models and the optimal model is bounded in Proposition \ref{prop:thm8}, and the maximum difference between an honest model and the optimal model is bounded in Proposition \ref{prop:lemma3} with a high probability, which are exactly the third and the first term respectively in Equation \ref{eqn:expansion}.

Now let us consider \(\frac{1}{\sum^K_{j=1}z^{(j)}}\sum_{i \in \mathcal{B}}z^{(i)} \left|\hat{y}^{(i)} -\tilde{y} \right| \). For an attacker \(a \in \mathcal{B}\), there are two cases:
\begin{itemize}
     \item \(\left|e^a\right| \leq \frac{\sqrt{2}\lambda}{\sqrt{K}}\). 
    From Equation \ref{eqn:normalized_residual} and \(\left|e^a\right| \leq \frac{\sqrt{2}\lambda}{\sqrt{K}}\), we have
$\left|\hat{y}^{(a)} -\tilde{y} \right| \leq \frac{C}{\sqrt{K}}$, where $C = \gamma\sqrt{2}\lambda\widetilde{|r|}(1 + \frac{5}{K-1})$.

\item  \(\left|e^a\right| > \frac{\sqrt{2}\lambda}{\sqrt{K}}\). If \(\left|e^a\right| > \left|\frac{\sqrt{2}\lambda}{\sqrt{K}\delta}\right| \) then it can be eliminated from Equation \ref{eqn:confidence}. 
From Equation \ref{eqn:normalized_residual} and \(\left|e^a\right| > \frac{\sqrt{2}\lambda}{\sqrt{K}}\), we have
$\frac{C}{\sqrt{K}} < \left|\hat{y}^{(a)} -\tilde{y} \right| \leq \frac{C}{\sqrt{K}\delta}$.
\end{itemize}

Combine two cases, we know that \(\left|\hat{y}^{(a)} -\tilde{y} \right| \leq \frac{C}{\sqrt{K}\delta}\), and we  have the following inequality,
\begin{equation}
\frac{1}{\sum^K_{j=1}z^{(j)}}\sum_{a \in \mathcal{B}}z^{(a)} \left|\hat{y}^{(a)} -\tilde{y} \right| 
\leq \sup_{a \in \mathcal{B}}\left|\hat{y}^{(a)} -\tilde{y} \right| 
\leq \frac{C}{\sqrt{K}\delta}.
\end{equation}
Therefore, we prove that
\begin{equation}
\left| \frac{1}{\sum^K_{j=1}z^{(j)}}\sum^K_{i=1}z^{(i)} \hat{y}^{(i)} - \mu \right| = \Tilde{\mathcal{O}}(\frac{1}{\sqrt{S}} + \frac{1}{\sqrt{SK}} + \frac{1}{S} + \frac{1}{\sqrt{K}\delta}).
\end{equation}
\end{proof}

\section{Conclusion}
Federated learning utilizes private data on multiple devices to train a global model, but the simple aggregation algorithm in federated learning is vulnerable to malicious attacks. To tackle this problem, we present a novel aggregation algorithm with residual reweighting. Our experiments on computer vision, natural language processing, and financial data show that our approach is robust to label-flipping and backdoor attacks while prior aggregation methods are not. Our algorithm is easy to implement and readily incorporated into existing federated learning frameworks. We hope our proposed aggregation algorithm can make federated learning more practical and robust in the future.

\newpage
\clearpage
\bibliography{aaai21}

\end{document}

%% file: math_commands.tex
\usepackage{amsmath,amsfonts,bm}

\def\eqref#1{equation~\ref{#1}}

\def\1{\bm{1}}

\DeclareMathAlphabet{\mathsfit}{\encodingdefault}{\sfdefault}{m}{sl}
\SetMathAlphabet{\mathsfit}{bold}{\encodingdefault}{\sfdefault}{bx}{n}

